\documentclass[10pt, conference, compsocconf]{IEEEtran}
\ifCLASSINFOpdf
  \usepackage[pdftex]{graphicx}
  \DeclareGraphicsExtensions{.pdf,.jpeg,.png}
\else
  \usepackage[dvips]{graphicx}
  \DeclareGraphicsExtensions{.eps}
\fi
\usepackage[cmex10]{amsmath}
\usepackage{algorithmic}
\usepackage{array}
\usepackage{mdwmath}
\usepackage{mdwtab}
\usepackage{xcolor}

\begin{document}
\title{Towards Robust Human Activity Recognition from RGB Video Stream with Limited Labeled Data \vspace{-1.3em}}


\author{\IEEEauthorblockN{Krishanu Sarker}
\IEEEauthorblockA{Dept. of Computer Science\\
Georgia State University\\
Atlanta, GA, USA\\
ksarker1@student.gsu.edu}
\and
\IEEEauthorblockN{Mohamed Masoud}
\IEEEauthorblockA{Dept. of Computer Science\\
Georgia State University\\
Atlanta, GA, USA\\
mmasoud1@student.gsu.edu}
\and
\IEEEauthorblockN{Saeid Belkasim}
\IEEEauthorblockA{Dept. of Computer Science\\
Georgia State University\\
Atlanta, GA, USA\\
sbelkasim@gsu.edu}
\and
\IEEEauthorblockN{Shihao Ji}
\IEEEauthorblockA{Dept. of Computer Science\\
Georgia State University\\
Atlanta, GA, USA\\
sji@gsu.edu}
}

%Telephone: (800) 555--1212, Fax: (888) 555--1212}


\maketitle
\vspace{-10mm}
\begin{abstract}
Human activity recognition based on video streams has received numerous attentions in recent years. Due to lack of depth information, RGB video based activity recognition performs poorly compared to RGB-D video based solutions. On the other hand, acquiring depth information, inertia etc. is costly and requires special equipment, whereas RGB video streams are available in ordinary cameras. Hence, our goal is to investigate whether similar or even higher accuracy can be achieved with RGB-only modality. In this regard, we propose a novel framework that couples skeleton data extracted from RGB video and deep Bidirectional Long Short Term Memory (BLSTM) model for activity recognition. A big challenge of training such a deep network is the limited training data, and exploring RGB-only stream significantly exaggerates the difficulty. We therefore propose a set of algorithmic techniques to train this model effectively, e.g., data augmentation, dynamic frame dropout and gradient injection. The experiments demonstrate that our RGB-only solution surpasses the state-of-the-art approaches that all exploit RGB-D video streams by a notable margin. This makes our solution widely deployable with ordinary cameras. 
\end{abstract}


% \fi
\IEEEpeerreviewmaketitle

\vspace{-2mm}
\section{Introduction}
\vspace{-.75mm}
Human action is inherently complex due to the inter-class affinity and intra-class diversity. Recognizing activity is hence a difficult task, which has attracted numerous researchers' attention \cite{cheng2015advances, herath2017going,chen2017survey}. Even though state-of-the-art image classification methods have surpassed human level accuracy \cite{russakovsky2015imagenet}, performance of methods proposed in the literature for activity recognition/classification is still unsatisfactory, especially methods based on RGB video stream \cite{cheng2015advances}. 

Despite many efforts, human action recognition from RGB video streams still lacks in accuracy compared to the progresses made in multi-modal data that includes depth enabled RGB-D video. One of the reasons behind this is that multi-modal datasets provide higher quantity of information, and the extracted depth information give precise detection of movement in the scene. However, depth enabled cameras are expensive and require special settings for many possible use-cases  of human action recognition. The economical factor and the installation complexity are the main reasons that most of the surveillance systems are using RGB cameras. Therefore, focusing our attention on the more popular RGB videos for detecting and classifying human motion would benefit the users and the real life applications.

Traditionally, the works on RGB video stream are based on handcrafted features \cite{schuldt2004kth,zhang2008motion,liu2008learning,bregonzio2009recognising,wang2013dense}. These approaches are highly data dependent. Due to this problem, these methods are very brittle and hard to deploy in real life in spite of higher accuracy they achieve. With the advent of deep learning, methods were proposed where features could be automatically extracted \cite{baccouche2011sequential, ji20133d, simonyan2014two, masoud2017automatically}. Successful use of deep learning with image classification inspired researchers to deploy such methods in video classification \cite{herath2017going}. These methods use raw RGB frames, often coupled with motion, to learn the temporal features. Even though these methods automate the feature extraction task, they often struggle to gain high performance due to complex background and partial occlusion of subjects in video streams. Hence, more robust, automated action recognition system is yet to be developed.

Approaches based on multiple modalities of data \cite{chen2015utd,wang2016action,li2017joint,rahmani2017learning}, however achieves higher accuracy even with complex actions. In these approaches, skeleton information extracted from depth images are proven very efficient in extracting important features of action. Inspired by this, in the paper we propose to use a technique that aims at separating salient features from the scene by extracting skeleton key-points from RGB-only video streams. This is a distinct departure from all previous approaches that either use raw RGB video stream as input directly or use skeleton key-points extracted from the depth information for activity recognition. Specifically, we utilize Openpose API \cite{cao2017openpose} as a black box to extract the skeleton key-points from each RGB frame. These key-point features are then fed into a Bidirectional Long Short Term Memory (BLSTM) based model to learn the spatio-temporal representations, which are subsequently classified by a softmax classifier. 

We use RGB-only modality for our experimental evaluations whereas state-of-the-art methods utilized multiple available modalities (depth, inertia and skeleton data). This essentially reduces training data to one forth for our experiments. Hence, we are dealing with one of the key challenges of deep learning, i.e., training with limited labeled data. To train the deep network effectively, we explore data augmentation and a few algorithmic approaches. Experiments on two popular and challenging benchmarks validate the effectiveness of these techniques and our RGB-only solution even surpasses the state-of-the-arts approaches that all exploit RGB-D videos. We believe that the proposed RGB-only scheme is more cost effective and highly competitive than RGB-D based solutions and therefore widely deployable.

Our key contributions are summarized in the following:
\begin{itemize}
\item  There exist previous methods in literature that are either based on skeleton extracted from depth data or purely based on raw RGB data for human activity recognition. To the best of our knowledge, we are the first to leverage skeleton key-points extracted from RGB-only videos for human activity recognition. 
\item We leverage data augmentation to tackle the problem of limited labeled data in deep learning, and compensate the data sparsity issue caused by using RGB-only modality. 
%We propose a novel deep classifier based on BLSTM. We also propose two design choices on baseline model; Dynamic Frame Dropout, and Gradient Injection.

\item Additionally, we explore a few algorithmic approaches such as Dynamic Frame Dropout and Gradient Injection to effectively train the deep architecture. 

\item We evaluate our proposed framework on two popular and challenging benchmarks, and demonstrate for the first time that using RGB-only streams we can surpass the state-of-the-art RGB-D based solutions, and make our RGB-only solution widely deployable. 
\end{itemize}

The rest of the paper is organized as follows. Related works are discussed in Section~\ref{sec:2}. We present our proposed architecture in Section~\ref{sec:3} and its effective training in Section~\ref{sec:4}. Experimental results with comparison to the state-of-the-arts are presented in Section~\ref{sec:5}, followed by a discussion and future works in Section~\ref{sec:6}. 
\vspace{-1mm}
\section{Related Works} \label{sec:2}
\vspace{-1.5mm}
Human activity recognition has been extensively studied in the recent years \cite{cheng2015advances,herath2017going,chen2017survey}. Most of state-of-the-art methods exact handcrafted features from RGB videos and rely on traditional shallow classifiers for activity classification~\cite{schuldt2004kth,zhang2008motion,liu2008learning,bregonzio2009recognising,wang2013dense}. For example, Schuldt et al.~\cite{schuldt2004kth} present a method that identifies spatio-temporal interest points and classifies action by using SVMs. Zhang et al. \cite{zhang2008motion} introduce the concept of motion context to capture spatio-temporal structure. Liu and Shah \cite{liu2008learning} consider the correlation among features. Bregonzio et al. \cite{bregonzio2009recognising} propose to calculate the difference between subsequent frames to estimate the focus of attention. These methods often achieve very high accuracy. However, since handcrafted features are highly data dependent, these methods are not very robust to the change of environments. We instead utilize OpenPose to extract the salient skeleton features from raw RGB frames, which makes the proposed method less data dependent, robust to different environments and therefore widely deployable in real life applications. 

Deep learning based approaches for human activity recogntion have also been explored extensively~\cite{baccouche2011sequential,ji20133d,simonyan2014two}. For example, Baccouche et al. \cite{baccouche2011sequential} propose to use Convolutional Neural Network (CNN) to extract spatial features and then use LSTM to learn the temporal features. Ji et al. \cite{ji20133d} present 3D CNN to classify actions which learns inherent temporal structure among the consecutive frames. A two-stream CNN based method is proposed in \cite{simonyan2014two}. In contrast to state-of-the-art handcrafted feature design approaches, deep learning based approaches use an end-to-end learning pipeline and extract feature representations automatically from data. However, these methods often fail to achieve higher accuracy as the high level features extracted from CNN are blurry and incapable of capturing the sharp changes in video streams. This is primarily because convolution and pooling tries to accurately capture the overall structure, while repetitive convolution and pooling operations often ignore the fine-grained details. 

In order to solve the aforementioned issues, skeleton information from RGB-D video has been widely studied to improve recognition accuracy \cite{wang2016action, li2017joint, rahmani2017learning, du2015hierarchical, liu2017enhanced}. Observations from seminal work by Johansson \cite{Johansson1973} suggests that a few movement of human joints is sufficient to recognize an action. Recently, Liu et al. \cite{liu2017enhanced} propose a CNN based approach leveraging the skeleton data. In \cite{du2015hierarchical} the authors propose hierarchical bidirectional Recurrent Neural Network (RNN) to classify the human actions. Methods proposed in \cite{hou2016skeleton} and \cite{wang2016action} utilize skeleton data on three CNN streams that are pretrained on large ImageNet Dataset \cite{deng2009imagenet}. Li et al. \cite{li2017joint} use view invariant features from skeleton data to improve over \cite{hou2016skeleton} and \cite{wang2016action}, and they used similar four stream pretrained models. All these methods utilize skeleton data, either extracted from depth data or Kinect. Inspired from these works, we adopt a bidirectional LSTM in our method; instead of extracting skeleton data from depth information as in other methods, we extract skeleton keypoints from RGB frames, which are available in ordinary digital cameras. 

%Evidently, methods leveraging skeleton data, extracted from depth information, edge over methods that simply take raw frames as input. 
In addition, there exist a few CNN and LSTM based approaches for activity recognition from RGB-only data \cite{donahue2015long,yue2015beyond}. However, none of them pay special attention to the issue of training deep network effectively on limited labeled data. We emphasize more on algorithmic approaches to address the training issues of deep networks with limited training data to alleviate  overfitting and gradient vanishing problems. Enhanced by these techniques, our RGB-only solution is able to surpass the state-of-the-arts that all exploit RGB-D streams.
\vspace{-2mm}
\section{Methodology}\label{sec:3}
\vspace{-2mm}
In this section, we present an end-to-end framework for human activity recognition from RGB video containing human silhouette. To make our discussion self-contained, we review some important concepts in the following subsections.
\begin{figure}[tbp]
\centerline{\includegraphics[width=3.2in]{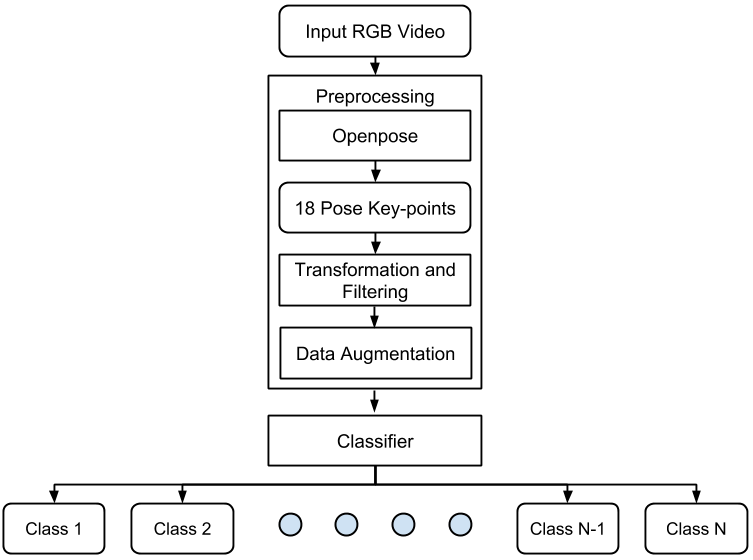}}
\caption{Overview of proposed method.}
\vspace*{-7mm}
\label{fig:pipeline}

\end{figure}
\subsection{Overview}
\vspace{-1mm}
Our proposed architecture aims to classify human actions from RGB-only streams to make our approach most amenable to ordinary cameras. We formulate our problem as learning the mapping, $\textbf{F}: x \rightarrow \ell$, where $x$ is the raw video and $\ell$ is the collection of action categories. After training, $\textbf{F}$ is used to classify the test samples.

Fig.~\ref{fig:pipeline} shows the overall pipeline of the proposed method. First, we extract pose key-points of human silhouette from input raw RGB video using OpenPose API~\cite{cao2017openpose}. We then preprocess the extracted pose key-points to improve the quality of the feature representations. After preprocessing, we use a variety of data augmentation techniques on the extracted keypoints to increase the training data size (and therefore mitigate the problem of data scarcity). In the end, the augmented training set is used to train our classifier. We used deep BLSTM \cite{hochreiter1997lstm} network coupled with MLP \cite{sarle1994neural} as our classifier. Overfitting is a major drawback for LSTM when dealing with small dataset. In addition to data augmentation, we therefore explored additional regularization techniques, such as dropout and L2 regularization to  prevent our model from overfitting. We also propose Dynamic Frame Dropout to reduce the redundant frames from a video and improve the robustness of the BLSTM classifier. To mitigate the vanishing gradient issue of LSTM, we introduce Gradient Injection to improve gradient flow. We will discuss each of these components in greater details in the following subsections. 

\subsection{OpenPose}
\vspace{-1mm}
OpenPose \cite{cao2017openpose} is an open source API that can be used to detect the 2D poses of multiple human subjects in an image. The API leverages a novel two stream multi-stage CNN, which facilitates it to work in real time. The methodology proposed in \cite{cao2017openpose} was ranked number one in COCO 2016 keypoints challenge. The input of the architecture is raw RGB image and the output is 15 or 18  pose key-points along with the part joining edges. More details about the architecture and working principle can be found in \cite{cao2017openpose}. In our work, we treat OpenPose as a black box with raw video frames as inputs and 18 pose key-points per person as output. %Fig.~\ref{fig:render} shows an example of OpenPose output, where 15 key-points were detected by OpenPose with higher confidence and remaining three low confidence key-points were excluded. 
%
%\begin{figure}[tbp]
%\centerline{\includegraphics[width=0.5in]{icmi_rendered}}
%\caption{Output of OpenPose: Rendered pose on silhouette.}
%\label{fig:render}
%\end{figure}
\subsection{LSTM}
\vspace{-1mm}
Long Short-Term Memory (LSTM) \cite{hochreiter1997lstm} is a descendant of Recurrent Neural Network (RNN) especially designed to adapt long range dependencies when modeling sequential data. RNN, in general, has been proven very successful in modeling sequences that have strong temporal dependency. However, vanishing gradient problem makes Vanilla RNN hard to train \cite{bengio1994lstmvanish}. LSTM migitates this issue by introducing non-linear gates regulating the information flow. In addition, vanilla LSTM can only learn from past contexts, whereas Bidirectional LSTM (BLSTM)~\cite{graves2005bidirectional} can be used to learn both from past and from future context by utilizing forward and backward layers. For human activity recognition task, we found that BLSTM is a more suitable architecture than vanilla LSTM as incorporating long term dependency in both directions in general helps improve learning of sequential data.
%
%\begin{figure}[t]
%\centerline{\includegraphics[width= 3 in]{brnn.png}}
%\caption{Bidirectional RNN, redrawn from \cite{graves2013brnn}}
%\label{fig:brnn}
%\end{figure}
\vspace{-1mm}
\subsection{Preprocessing} 
\vspace{-1mm}
The preprocessing step represents the first step of our end-to-end pipeline where the raw video frames are fed into the OpenPose API. The output of OpenPose for each video frame is a matrix of shape $(n_{pose},(a,b),c)$. Here, $n_{pose}$ is the number of pose key-points, $(a,b)$ is the coordinates of the key-points in Cartesian plane and $c$ is the confidence score of the respective key-point. To simplify our problem, we put a constraint that each frame can contain at most one person, and hence the value of $n_{pose}$ here is 18. When all pose key-points are extracted from a video, we use a filter to set the pose keypoints values that has confidence lower than a threshold value, $\Theta$, to zero. Later, we mask these zero valued keypoints in order to avoid learning from these points. Afterwards, the pose matrix is flattened and converted into a vector, $\Lambda$, of size $n_{pose}*2$, excluding the confidence value. We concatenate each pose frame into a 2 dimensional matrix of shape $(n_{frame},v)$, where $n_{frame}$ is the number of frames in the video and $v$ is the length of pose vector, $\Lambda$. 
\subsection{Proposed Network Architecture}
\vspace{-1mm}
Our proposed deep architecture combines deep BLSTM layers and MLP. We use five consecutive BLSTM layers with dropout to regularize the model training. We utilize Batch Normalization (BN) after each BLSTM layer to keep the data normalized throughout the pipeline. We feed the output of the Deep BLSTM layers to the MLP consisting of two Dense layers. For intermediate hidden BLSTM and Dense layers, we have utilized the Parametric Rectified Linear Unit (PReLU) \cite{he2015prelu} activation layer. We use the softmax function for the final output layer to produce probabilistic score for each class. Categorical cross-entropy is used to measure the loss of our proposed network. We utilized RMSprop optimizer \cite{tieleman2012rmsprop} to minimize the loss function. %If $f'(\theta_t)$ is the first order derivative of loss function with respect to the parameters at time $t$, given a step rate $\alpha$ and a decay term $\rho$, RMSprop is updated as shown in \eqref{eq:7} \eqref{eq:8} and \eqref{eq:9}. 
%\begin{equation} \label{eq:7}
%r_t = (1 - \rho)f'(\theta_t)^2 + \rho r_{t-1}
%\end{equation}
%\begin{equation} \label{eq:8}
%v_{t+1} = \dfrac{\alpha}{\sqrt{r_t}}f'(\theta_t)
%\end{equation}
%\begin{equation} \label{eq:9}
%\theta_{t+1} = \theta_{t} - v_{t+1}
%\end{equation}
%
%

\section{Effective Training of BLSTM}\label{sec:4}
\begin{figure}[tbp]
\centerline{\includegraphics[width=3.2in]{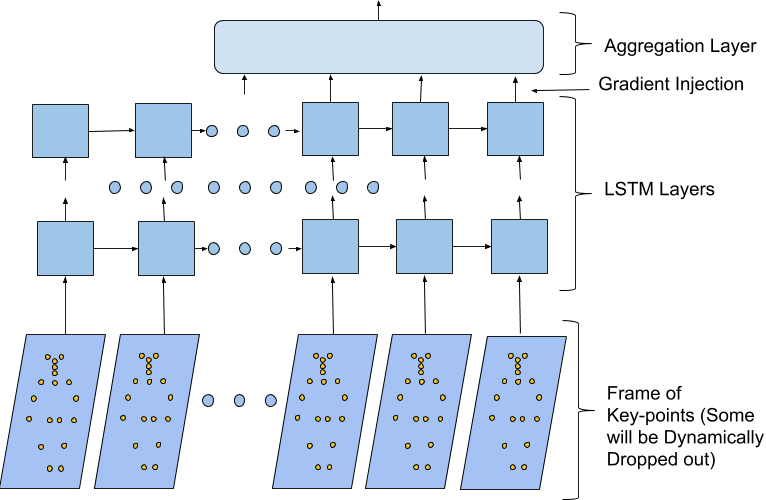}}
\caption{BLSTM with Dynamic Frame Dropout and Gradient Injection.}
\label{fig:high}
\vspace*{-6mm}
\end{figure}

\vspace{-1mm}
It's challenging to train the BLSTM architecture as the number of parameters can be easily larger than several hundred millions, while the number of training data for the purpose of model parameter estimation is typically very small (e.g., 2-3 orders of magnitude lower). Therefore, special attention is needed to address the effective training of BLSTM, otherwise overfitting, gradient vanishing can quickly plague the learning process. In the following, we discuss a few techniques we explored to train the deep model effectively, with the high level scheme demonstrated in Fig.~\ref{fig:high}.

\vspace{-1mm}
\subsection{Dynamic Frame Dropout}
\vspace{-1mm}
We propose to utilize Dynamic Frame Dropout (DFD) to reduce data redundancy. As different actions require different time span, and often there are redundant information in consecutive frames, taking all frames into account actually occludes crucial information and hampers the learning. Techniques like randomly dropping frames or dropping each $n$ frames etc. are often used in state-of-the-art methods. However, doing so naively may result in loss of important information. Instead of randomly dropping frames, DFD drops frames only if they contain information that are almost redundant to their preceding frames. This not only reduces the computation complexity but also introduces stochasticity and help to regularize the model training in a similar spirit of dropout.

Specifically, we measure the redundancy between two consecutive frames by computing the euclidean distance between their feature vectors, i.e., pose key-points of two consecutive frames. The lower distance corresponds to similarity and the higher distance means these frames actually have meaningful differences. Empirically, we set a cutoff threshold, $\hat{c}=15$. If $d$ is distance between $frame_1$ and $frame_2$ and $d<\hat{c}$, then we drop $frame_2$. According to our experiments that follow, this setup of $\hat{c}$ drops 20 to 25 frames per video that carry information with minimal significance.
\vspace{-1mm}
\subsection{Gradient Injection}
\vspace{-1mm}
Although LSTM serves as the solution of vanilla RNN for gradient vanishing problem, it itself faces this issue in some degree when training deep model \cite{hsu2016exploiting}. LSTM many to one architecture is often used as the final layer of network for video classification. This creates a dependency on processing the whole video sequence before we can perform classification. However, a video can often be clearly classified before having to see all the frames till the end. Hence, to improve gradient flow and avoid gradient vanishing problem, we connect the MLP classification layer to the last $K$ time-steps and this allow the model to classify a video by incorporating multiple step information. When back-propagating error to update model parameters, this also allows the gradient to be propagated earlier in time and mitigate the gradient vanishing problem. We call this technique Gradient Injection (GI). In other words, we utilize many to many architecture of LSTM at the top layer to allow gradients flow from multiple time steps, consequently, reducing the problem of vanishing gradient. Moreover, as outputs from multiple time steps are now available, it creates an ensemble of multiple outputs and reduces dependency on all the video frames. 
%ectvi
\vspace{-1mm}
\subsection{Data Augmentation}
\vspace{-1mm}
%Our DCNN has millions of parameters, while only several thousand of training images are available. In order to reduce overfitting, we augment the training data by rotating the positive training images through 360. These images are also horizontally flipped to double the training images. This increases the number of training examples of body parts with different spatial relationships with its neighbors

Training a deep networks with limited amount of labeled training data is a major challenge in supervised learning paradigm. Our goal of achieving state-of-the-art performance with RGB-only data modality faces the same brick wall: insufficient training data. According to our problem formulation, we only leverage RGB data modality. Data augmentation has been proven very successful in supervised learning for image analysis. Inspired by this, we have explored several data augmentation techniques to solve the data scarcity problem. In our case, instead of the raw input video, we take skeleton key-point features as the input for data augmentation. We use translation, scaling and random noise to augment skeleton data. To keep the augmentation consistent throughout a single sample, we deploy same transformation on each key-point frames of that sample. 

In the experiments that follow we evaluate the significance of each of these techniques when training deep networks using limited training data. 
\vspace{-1mm}
\section{Experimental Results} \label{sec:5}
\vspace*{-1mm}
The primary goal of this paper is to show that by using RGB-only data modality with limited training data, we can achieve similar or higher accuracy on action recognition task than the state-of-the-arts that use RGB-D video streams. We have tested our proposed method with two widely used datasets, KTH \cite{schuldt2004kth} and UTD-MHAD~\cite{chen2015utd}. We focus on UTD-MHAD as this is a complex dataset offering multiple modalities and current state-of-the-art methods utilize data modalities consisting depth information to classify actions. Extensive experiments show that with the effective training techniques discussed in Section~\ref{sec:4}, i.e., data augmentation, dynamic frame dropout and gradient injection, the proposed RGB-only solution surpasses the state-of-the-art methods by a notable margin. On RGB-only dataset such as KTH, our method outperforms all the other methods reported in the literature, demonstrating the versatility of the proposed architecture and techniques for human activity recognition.

We implemented our system in Python with Tensorflow backend on a GPU cluster with Intel Xeon CPU E5-2667 v4 @ 3.20GHz with 504 GB of RAM and NVIDIA TITAN Xp with 12 GB of RAM and 3840 cuda cores. In our experiments, we empirically set learning rate, $lr = 5e^{-5}$ for RMSprop optimizer. We report confidence interval based on 50 bootstrap trials. More details about datasets we evaluated our model on and comparative experimental studies with state-of-the-art literatures are presented next. 
%
%\begin{figure}[tbp]
%\begin{subfigure}[b]{0.5\textwidth}
%                \centering
%                \includegraphics[width=.9\linewidth]{kth}
%        \end{subfigure}%
%        
%        
%        \begin{subfigure}[b]{0.5\textwidth}
 %               \centering
%                \includegraphics[width=.9\linewidth]{utd_mhad}
%        \end{subfigure}%
%        \caption{Sample frames from KTH (top row) and UTD-MHAD (bottom row) datasets. }
%\label{fig:kth}
%\end{figure}
\vspace{-1mm}
\subsection{Dataset}
\vspace{-1mm}
KTH \cite{schuldt2004kth} is an RGB-only video dataset containing six action classes (walking, running, boxing, hand-waving, and hand-clapping), performed by 25 subjects in various conditions. KTH dataset provides full silhouette figure in all the sequences, which satisfies our requirements of pose-based activity recogntion\footnote{To extract pose key-points reliably, we need the subjects in video streams expose their full silhouettes.}. We have followed the same experimental setup stated in \cite{schuldt2004kth}. 

UTD-MHAD \cite{chen2015utd} is a multi-modal action dataset containing 27 actions performed by 8 subjects (4 males and 4 females) performing same action 4 times, a total 861 sequences. This dataset provides four temporally synchronized data modalities; RGB videos, depth videos, skeleton positions, and inertial signals from Kinect camera and a wearable inertial sensor. We follow 50-50 train-test split similar to \cite{chen2015utd}. In the experiments we only use the RGB modality to evaluate our proposed method. 
\iffalse
%%%%%%%%%%%%
\begin{table}[h]
\centering
\caption{UTD-MHAD dataset actions}
\label{tbl:utd_act}
\begin{tabular}{llll}
\hline
Index & Action        & Index & Action      \\ \hline
1  & swipe\_left       & 15 & tennis\_swing \\
2  & swipe\_right      & 16 & arm\_curl     \\
3  & wave              & 17 & tennis\_serve \\
4  & clap              & 18 & push          \\
5  & throw             & 19 & knock         \\
6  & arm\_cross        & 20 & catch         \\
7  & basketball\_shoot & 21 & pickup\_throw \\
8  & draw\_x           & 22 & jog           \\
9  & draw\_circle\_CW  & 23 & walk          \\
10 & draw\_circle\_CCW & 24 & sit2stand     \\
11 & draw\_triangle    & 25 & stand2sit     \\
12 & bowling           & 26 & lunge         \\
13 & boxing            & 27 & squat         \\
14 & baseball\_swing   &    &              
\end{tabular}
\end{table}
%%%%%%%%%%%%%%%%%
\fi

%

\begin{figure}[tbp]
\centerline{\includegraphics[width=3in]{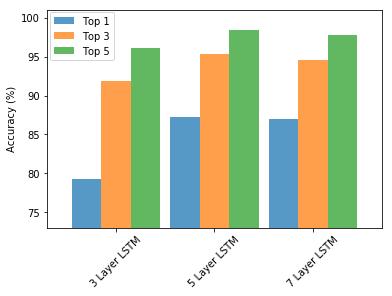}}
\caption{Accuracy comparison on the UTD-MHAD dataset on our models with different number of LSTM layers.}
\label{fig:layer}
\vspace*{-6mm}
\end{figure}

\vspace{-1mm}
\subsection{Experiments on UTD-MHAD dataset}
\vspace{-1mm}
We first explore the choices of depth of the network. We test our baseline BLSTM model with three settings: 3-Layer, 5-Layer and 7-Layer models. Fig. \ref{fig:layer} presents the accuracy of these models on top 1, 3 and 5 categories. Evidently, the 5 and 7 layer models outperform the 3 layer model, and the performances of the 5 and 7 layers models are almost on par. Therefore, the 5-layer model reaches a good accuracy and model complexity balance. We use the 5-layer model architecture in all our following experiments. Notice that the 3-layer model has shown comparative accuracy on top 3 and 5 categories with other two models, indicating that deeper models mainly boost the top-1 accuracy. 

\begin{figure}[bp]
\vspace*{-4mm}
\centerline{\includegraphics[width=3in]{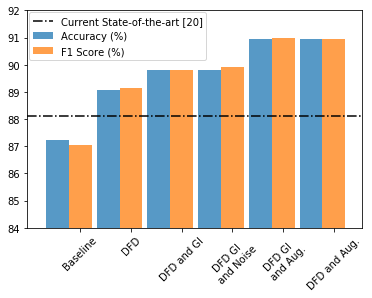}}
\caption{Accuracy comparison of the different design choices on the UTD-MHAD dataset.}
\label{fig:design_choice}
\vspace*{-6mm}
\end{figure}

To understand the impact of different techniques over the baseline BLSTM model (e.g., DFD, GI and data augmentation), we evaluate the accuracy of the BLSTM model as we enable them in a cumulative fashion. We begin our experiments on UTD-MHAD dataset using our baseline BLSTM model. Then the second model includes DFD; the third one includes both DFD and GI; in the fourth model we use random jittering to augment data, and finally in the fifth and last model we use affine transformation as data augmentation. Fig. \ref{fig:design_choice} shows the comparison among all these models on accuracy and F1 score. As we can see, the baseline BLSTM model itself reaches an accuracy of 87\%  (about 1\% behind the state-of-the-art accuracy 88\% \cite{li2017joint}); by including DFD on top of baseline, our method achieves an accuracy of 89.06\% outperforming state-of-the-art by 1\%. On top of that, GI and data augmentation further boost the performance to ~91\%. An interesting phenomena to observe here is that although GI does not have much effect on top of data augmentation, it helps gaining performance over DFD. In total, utilizing data augmentation we gain 2\% accuracy over DFD model, while using random data jittering is less effective and does not improve accuracy. 

To investigate the effect of data augmentation on the predictive accuracy, we experiment with incremental data augmentation. The results are summarized in Table \ref{tbl:utd_aug}. As can be seen, data augmentation regularizes the model training and helps model avoid overfitting. As a result, when we increase the training dataset by data augmentation, the top-1 error rate is consistently reduced. We also notice that data augmentation does not have much effect on top-3 error rate, indicating that data augmentation mainly boosts correct answers from top-3 positions to top-1 positions.
\vspace{-1mm}
\begin{table}[h]
\caption{Effect of Data Augmentation on the UTD-MHAD dataset.}
\vspace{-2mm}
\label{tbl:utd_aug}
\renewcommand{\arraystretch}{1.3}
\centering
\begin{tabular}{| c | c | c |} 

\hline
\textbf{Augment Size}  & \textbf{Top-1 Error (\%)}        & \textbf{Top-3 Error (\%)}    \\ 
\hline
\hline
0   &  10.94 & 4.01 \\
\hline
430  &  9.75  & 3.75 \\
\hline
860  &  9.35  & 3.68\\
\hline
1290 &  9.09  & 3.69\\
\hline
1720 &  9.05  & 3.66\\
\hline
\end{tabular}
\vspace{-2mm}
\end{table}

\begin{figure}[tbp]
\centerline{\includegraphics[width=2.9in]{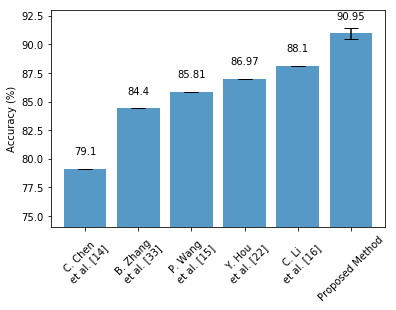}}
\caption{Accuracy comparison on the UTD-MHAD dataset. \cite{chen2015utd}, \cite{wang2016action},  \cite{li2017joint}, \cite{hou2016skeleton} and \cite{zhang2017action} use depth enabled modalities, while our method use RGB-only modality (confidence interval of our method is also included).}
\label{fig:utd_acc}
\vspace*{-6mm}
\end{figure}

Finally, we summerize the results of our proposed method with state-of-the-art methods \cite{chen2015utd,wang2016action,hou2016skeleton,li2017joint,zhang2017action} in Fig. \ref{fig:utd_acc}. Most of these methods use depth or inertia data modalities or both (section \ref{sec:2}). These data modalities are only available from depth enabled camera and provide more precise information of motions related to actions. On the contrary, we use RGB-only modality to train our model from scratch. It can be seen from Fig. \ref{fig:utd_acc} that our method achieves an accuracy of 90.95\% which outperforms all the state-of-the-art methods.

\vspace{-1mm}
\subsection{Experiments on KTH dataset}
\begin{figure}[tbp]
\centerline{\includegraphics[width=2.9in]{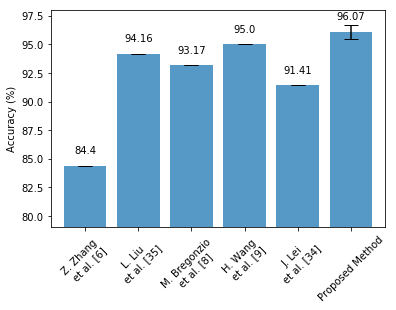}}
\caption{Accuracy comparison on KTH dataset with state-of-the-arts. \cite{lei2016continuous} use CNN based method, while \cite{zhang2008motion}, \cite{bregonzio2009recognising}, \cite{wang2013dense} and \cite{liu2016learning} utilize hand crafted features. (confidence interval of our method is also shown).}
\label{fig:kth_acc}
\vspace*{-6mm}
\end{figure}

\vspace{-1mm}
To further strengthen our hypothesis, we then compare our proposed method with the state-of-the-arts on the RGB only dataset, KTH \cite{schuldt2004kth}, with the results presented in Fig. \ref{fig:kth_acc}. We utilized similar training-testing split of the data as suggested in \cite{schuldt2004kth} to obtain the reported results. CNN based hybrid model proposed by Lei et al. \cite{lei2016continuous} achieves 91.41\% accuracy, which is outperformed by most of the state-of-the-art handcrafted feature based methods. On the other hand, \cite{wang2013dense, liu2016learning, zhang2008motion}, and \cite{bregonzio2009recognising} using handcrafted features achieve competitive accuracy. However, these methods are extremely data dependent; proposed handcrafted feature extractors in these methods cannot robustly work on heterogeneous data. Hence, these methods are not suitable for real world deployment. Our proposed method with data augmentation and dynamic frame dropout achieves 96.07\% accuracy, outperforming all the others. 

\vspace{-1mm}
\section{Conclusion and Future Work} \label{sec:6}
\vspace*{-1mm}
We propose an end-to-end framework that utilizes pose key-points extracted from OpenPose coupled with BLSTM for human activity recognition. A major difference to the state-of-the-art methods is that we use RGB-only modality while all the other methods use RGB-D modality. Effective training of deep networks in our setting is the major technical challenge as we typically have very limited training data and exploiting RGB-only modality exaggerates the difficulty even further. We therefore explore a number of algorithmic techniques like Dynamic Frame Dropout, Gradient Injection and Data Augmentation to train our framework effectively. Extensive experiments demonstrate the effectiveness of our BLSTM model and training methodologies, among which data augmentation is the most effective one. In the end, our RGB-only solution surpasses all the state-of-the-art methods that exploit RGB-D streams. This makes our solution cost effective and widely deployable with ordinary digital cameras. 

%The state-of-the-art methods achieve accuracy as high as 88.1\%, utilizing depth data modalities on the UTD-MHAD dataset. However, RGB is the only data-modality we leverage. Our proposed system successfully achieves significantly better performances with data augmentation. Moreover, using 5-fold cross validation (80-20 train-test split) we could achieve even better performance (above 94\%). This essentially proves that even with RGB-only data modality, data augmentation is sufficient to mitigate the problem of data sparsity and we were successful to train our model with augmented data and achieve better accuracy.

%Our proposed system successfully achieves significantly better performances on the datasets we evaluated with data augmentation. Moreover, using 5-fold cross validation we could achieve even better performance. This essentially proves that even with RGB-only data modality, it is possible to achieve better accuracy than the state-of-the-art methods that use all available modalities. Only reason these methods achieve higher accuracy is that, they have more data than us. We also show that data augmentation is sufficient to mitigate this problem and we were successful to train our model with only half data the other methods used and achieved better accuracy. 

Our experiments were conducted on the KTH and UTD-MHAD datasets, where there is only one person present per action and whole silhouette is visible. %Additionally, these datasets were collected in a more controlled environment which makes them less realistic. 
In the future, we would like to extend our method for multi-person datasets where some body parts can be partially occluded, which happen more often in real video surveillance applications.

\vspace{-1mm}
\section*{Acknowledgment}
\vspace{-1mm}
The authors would gratefully acknowledge the support of NVIDIA Corporation with the donation of the Titan Xp GPU used for this research.



% http://www.ctan.org/tex-archive/biblio/bibtex/contrib/doc/https://www.overleaf.com/project/5ba04cbea67f96296764d7da
%\bibliographystyle{IEEEtran}

{
\renewcommand{\baselinestretch}{0.70}
\bibliographystyle{IEEEtran}
\small

\bibliography{egbib}
}

\end{document}